\documentclass[pdflatex,sn-mathphys-num]{sn-jnl}

\usepackage{graphicx}%
\usepackage{multirow}%
\usepackage{amsmath,amssymb,amsfonts}%
\usepackage{amsthm}%
\usepackage{mathrsfs}%
\usepackage[title]{appendix}%
\usepackage{xcolor}%
\usepackage{textcomp}%
\usepackage{manyfoot}%
\usepackage{booktabs}%
\usepackage{algorithm}%
\usepackage{algorithmicx}%
\usepackage{algpseudocode}%
\usepackage{listings}%
\usepackage{multicol}

\theoremstyle{thmstyleone}%
\theoremstyle{thmstyletwo}%

\theoremstyle{thmstylethree}%

\begin{document}

\title[Article Title]{ECMNet:Lightweight Semantic Segmentation with Efficient CNN-Mamba Network}


\author[1,2]{\fnm{Feixiang} \sur{Du}}

\author[1]{\fnm{Shengkun} \sur{Wu}}

\affil*[1]{\orgdiv{TLU}}

\affil[2]{\orgdiv{SUT}}


\abstract{In the past decade, Convolutional Neural Networks (CNNs) and Transformers have achieved wide applicaiton in semantic segmentation tasks. Although CNNs with Transformer models greatly improve performance, the global context modeling remains inadequate. Recently, Mamba achieved great potential in vision tasks, showing its advantages in modeling long-range dependency. In this paper, we propose a lightweight \textbf{E}fficient \textbf{C}NN-\textbf{M}amba \textbf{Net}work for semantic segmentation, dubbed as \textbf{ECMNet}.  
ECMNet combines CNN with Mamba skillfully in a capsule-based framework to address their complementary weaknesses.
Specifically, We design a Enhanced Dual-Attention Block (EDAB) for lightweight bottleneck. In order to improve the representations ability of feature, We devise a Multi-Scale Attention Unit (MSAU) to integrate multi-scale feature aggregation, spatial aggregation and channel aggregation. Moreover, a Mamba enhanced Feature Fusion Module (FFM) merges diverse level feature, significantly enhancing segmented accuracy. Extensive experiments on two representative datasets demonstrate that the proposed model excels in accuracy and
efficiency balance, achieving 70.6\% mIoU on Cityscapes and 73.6\% mIoU on CamVid test datasets, with 0.87M parameters and 8.27G FLOPs on a single RTX 3090 GPU platform. 
}

\keywords{Semantic segmentation, Lightweight, Convolutional neural network, Mamba}



\maketitle

\section{Introduction}\label{sec1}

Semantic segmentation aims to assign a label to each pixel in a given image, which is widely applied in autonomous driving\cite{sanchez2025cola}, remote sensing\cite{jing2025hypergraph}, and agriculture\cite{luo2024semantic}, and more.

Early semantic segmentation primarily relied on CNNs, employing techniques like large convolutional kernels \cite{peng2017large}, dilated convolutions\cite{chen2017deeplab}, and feature pyramids\cite{zhao2017pyramid}  to extend receptive fields. 
However, these CNN-based approaches remained limited in capturing long-range dependencies.
The advent of Transformers\cite{vaswani2017attention} enabled more effective global context modeling in subsequent segmentation methods.
Learning global context dependencies is essential for extracting global semantic features, particularly in intensive tasks like semantic segmentation.
The rise of Visual Transformer (ViT)\cite{dosovitskiy2020image} has injected a new paradigm for semantic segmentation.
SETR\cite{zheng2021rethinking} slices images into sequences for the first time and captures global context feature through a self-attentive mechanism, outperforming traditional CNN models on complex scene datasets such as Cityscapes.
Meanwhile, SegFormer\cite{xie2021segformer} further optimized the architectural design by proposing a hierarchical Transformer encoder with a lightweight MLP decoder to achieve multi-scale feature fusion.
However, the square-level computational complexity of Transformer limits its application to high-resolution images with insufficient sensitivity to local details.

To tackle the limitation of the above single model and extract fine spatial details, some models treated semantic segmentation tasks by integrating CNN with Transformer.
For instance, HResFormer\cite{ren2025hresformer}, PFormer\cite{gao2025pformer},
and  DMFC-UFormer\cite{garbaz2025dmfc} have achieved satisfactory results in the field of
medical image segmentation. 
However, the self-attention mechanism in CNN-Transformer methods still poses challenges in terms of speed and memory usage when dealing with long-range visual dependencies, especially processing high-resolution images.

Unlike previous Transformer, Mamba\cite{gu2023mamba} shows great potential for high-resolution image by efficient sequence modeling with linear complexity.
Vision Mamba\cite{liu2024vmamba} have recently demonstrated remarkable success in various computer vision tasks.
For example, in the field of 3D medical imaging, SegMamba\cite{xing2024segmamba} achieves real-time inference on the colorectal cancer dataset CRC-500, with a speedup of 30\% compared to 3D UNet. 
In addition, CM-UNet\cite{liu2024cm} introduces a Mamba decoder into a CNN encoder to bridge local and global features through a channel-space attention mechanism, achieving higher mIoU on the ISPRS Vaihingen dataset.

To accommodate limited computational resources and mobile device application, lightweight semantic segmentation models receive higher attention. 
For example, LEDNet\cite{wang2019lednet} employed channel split-and-shuffle operations within residual blocks, significantly lowering computational complexity.
While CFPNet\cite{ding2024cfpnet} designed
Channel-wise Feature Pyramid (CFP) module to significantly reduces model parameters and model scale by extracting various level feature map and contextual feature information jointly.
LETNet\cite{xu2023lightweight} used an LDB module and FE module for enhanced efficiency and accuracy with reduced model complexity.

Motivated by the success of Mamba and lightweight approaches in semantic segmentation tasks, We propose ECMNet, an efficient CNN-Mamba hybrid network for lightweight semantic segmentation, optimized for minimizing model size and computational requirements.
The main contributions of our
paper are four folds:

\begin{itemize}
    \item We firstly propose a novel lightweight Efficient CNN-Mamba Network (EMCNet) for for semantic segmentation.
    ECMNet utilizes U-shape encoder-decoder structure as backbone and regards the Feature Fusion Module (FFM) as a capsule network to capture global context information. Specially, FFM introduces
    SS2D block, a variant of Mamba, to learn long-range dependencies. 
    \item We design a lightweight Enhanced Dual-Attention Block (EDAB) to extract multi-dimensional semantic information. EDAB consists of Dual-Direction Attention(DDA), Channel Attention (CA) and various convolution modules, realizing less model parameters and computational quantities.
    \item We develop a Multi-Scale Attention Unit (MSAU) to improve the representations ability of feature, which further refines the local details and global contextual information.
    \item ECMNet achieved 70.6\% mIoU on the Cityscapes dataset on the single RTX 3090 GPU with only 0.87M of parameters, realizing the better trade-off between performance and parameters. 
    Meanwhile, our proposed method achieved 73.6\% of the highest performance on the CamVid dataset, which demonstrates the effectiveness and generalization of our proposed ECMNet.
\end{itemize}

\section{Proposed Method}\label{sec3}
\subsection{Overall Network Architecture}\label{subsec2}
\begin{figure}[h]
    \centering
    \includegraphics[width=0.8\linewidth]{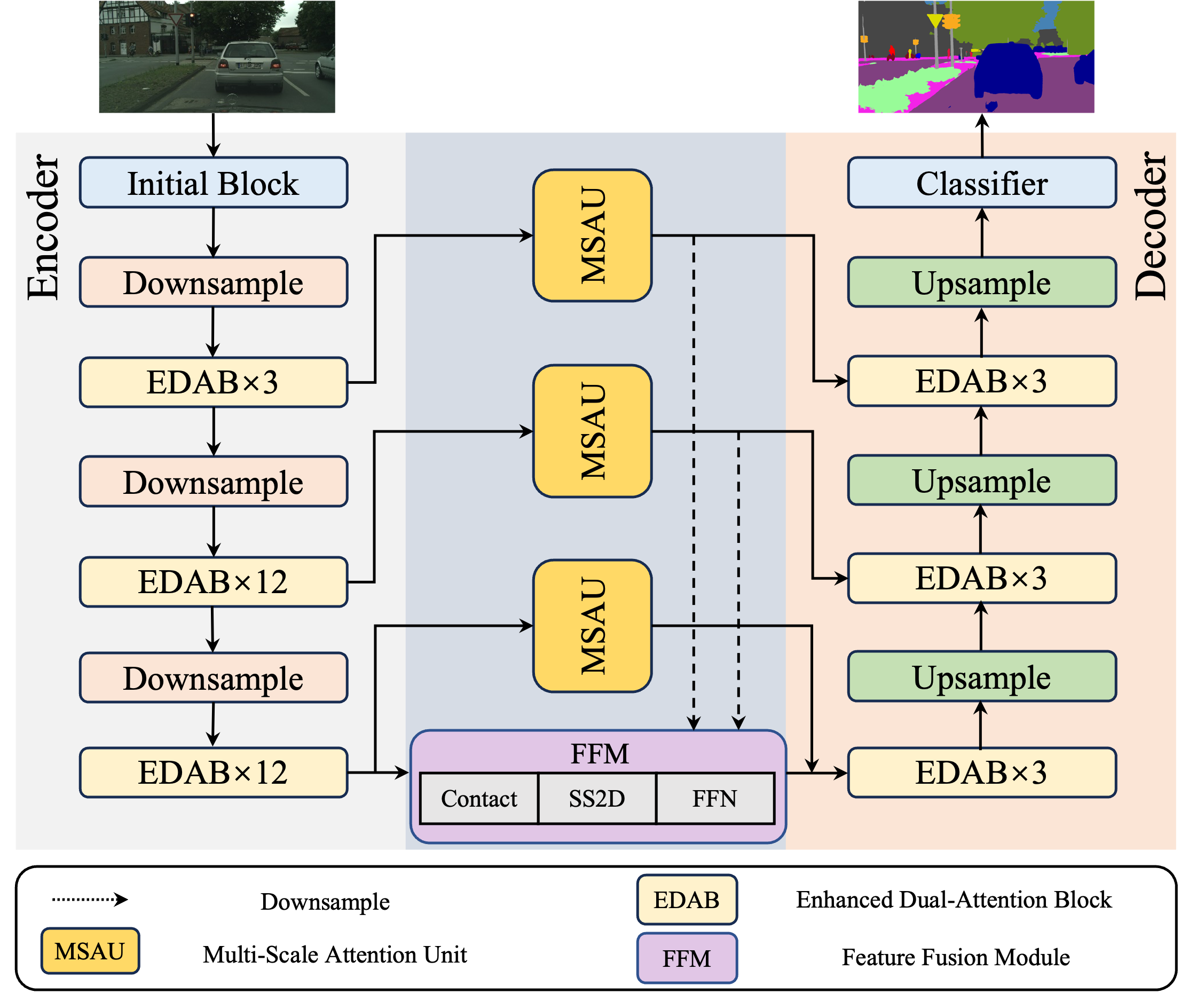}
    \caption{The overall network architecture of Efficient CNN-Mamba Network (EMCNet)}
    \label{fig:EMCNet}
\end{figure}

As shown in Figure \ref{fig:EMCNet}, the overall network architecture of our proposed EMCNet consists of four components: a CNN encoder improved with enhanced dual-attention blocks, a CNN decoder with subtle difference from encoder, an efficient Mamba-based feature fusion module and three long skip connections enhanced with multi-scale attention unit.
Specifically, the CNN-based encoder-decoder architecture extracts localized features for detailed spatial representation.
The Mamba-based FFM can capture complex spatial information and long-range feature dependencies by state space model (SSM) to optimize global feature representations and computational complexity. 
The three long-distance skip connections generate more high-quality segmentation by focusing on low-level spatial information and high-level semantic information respectively.
The above mentioned elaborated modules make it more efficient for ECMNet to fully integrate local and global feature information.

\subsection{Enhanced Dual-Attention Block}

The EDAB module is designed to focus different level feature information and keep network parameters as few as possible.
Firstly, the input feature passes through a bottleneck structure that utilizes a 1\texttimes 1 convolution to reduce the number of channels to half, significantly reducing the computational complexity and the number of parameters.
Obviously, this will sacrifice a part of the accuracy,it will be more beneficial to introduce 3\texttimes 1 convolution and 1\texttimes 3 convolution more than make up for the loss at this point.
Meanwhile, the two decomposed convolutions not only obtain a wider respective field for capturing a larger range of contextual feature information but also consider the model parameters and calculation complexity.
The core of EDAB lies in its two-branch path, which captures local and global feature information respectively.
Decompose convolution in one branch processes local and short-distance feature information, complemented by atrous convolution in the parallel branch for global feature integration.
Then, the channel contains most of the feature information and the spatial feature information is key to enhance performance and suppress noise interference.
Therefore, the two branches utilize Channel Attention (CA) and Dual-Direction Attention (DDA), which aims to build different attention matrix to learn multi-dimensional feature information and improve feature expression.
Finally, the outputs from both designed pathes and intermediate features are integrated and processed by a 1\texttimes1 point-wise convolution to restore the original channel dimensionality.
A channel shuffle strategy is applied at the end of EDBA to establish inter-channel correlations and overcome information fragmentation
The detail operation is shown as follows:
\begin{gather}
    F_{up\_branch} = Conv_{1\times3 }(Conv_{3\times×1} (Conv_{1\times1} (x))),\\
    F_{mid\_branch\_1} = Conv_{CA}(Conv_{1\times3,D} (Conv_{3\times1,D} (F_{up\_branch}))),\\
    F_{mid\_branch\_2} = Conv_{DDA}(Conv_{1\times3,D,R} (Conv_{3\times1,D,R} (F_{up\_branch}))),\\
    Y_1 = Conv_{CS}(f_{1\times1} (F_{up\_branch} + F_{mid\_branch\_1} + F_{mid\_branch\_2}) + x),
\end{gather}
where $x$ denotes the input of the EDAB, $Y_1$ denotes the output feature map of the EDAB, and $Conv_{k\times k}(\cdot)$ are normal convolution operation. 
Among the suffix, D denotes depth-wise convolution, R is the atrous rates of atrous convolution, CA represents Channel Attention, DDA represents Dual-Direction Attention and CS denotes the shuffle operation of channel.

\subsection{Multi-Scale Attention Unit}

On the one hand, lower layers preserve fine spatial details with limited semantics, on the other hand, higher layers offer strong semantic representation at lower spatial resolution.
Therefore, it is an efficient strategy to combine the low-level rich spatial information and high-level rich semantic information for semantic segmentation tasks.
Inspired by U-Net, we use the same resolution connections to integrate the high-level feature maps and low-level feature maps.
In order to better process the three long connections, we design a Multi-Scale Attention Unit (MSAU) to enhance the ability of feature representation.
The MSAU is carried out from two branches, one is the Multi-Scale Spatial Aggregation, the other is the Channel Aggregation.

In the Multi-Scale Spatial Aggregation, the input feature map is utilized 1\texttimes 1 convolution to convert from C channel to C/2 channel.
In order to reduces the amount of parameter and computation while retaining the ability of multi-scale feature extraction, the next feature map goes through different sizes of depth-separable convolution,such as 3\texttimes3, 5\texttimes5 and 7\texttimes7. 
Meanwhile, the outputs of different scale convolutions obtain multi-scale feature information enhancing the multi-scale perception capability of the model.
Then, the multi-scale fused feature map compresses the height dimension to 1 by adaptive average pooling, and generates a spatial attention map by 7\texttimes7 depth separable convolution, 1\texttimes1 convolution and Sigmoid activation function.
At the same time, by multiplying with the multi-scale fused feature map, the processed feature highlights the key spatial regions and suppresses the irrelevant information.
At last, the channel of model is converted from C/2 back to C by using 1x1 convolution, and the attention map reflects the importance of the different locations of feature map.
For channel aggregation, the input feature map uses average pooling and maximum pooling to obtain average channel features and maximum channel features respectively, which captures channel statistics from different angles.
The MSAU multiplies the spatial and channel aggregation results and adds them with the original input feature maps to obtain the output feature maps. 

This design allows the MSAU module to fused the low-level spatial information to the high-level semantic information more effectively, and further enhance the ability of feature expression
The detail operation can be defined as:
\begin{gather}
    X_1 = Conv_{(3\times3)}(Conv_{(1\times1)}(x)) + Conv_{(5\times5)}(Conv_{(1\times1)}(x)) \nonumber\\+ Conv_{(7\times7)}(Conv_{(1\times1)}(x)) \\
X_2 = Conv_{(1\times1)}(X_1 \otimes Sigmoid(Conv_{(7\times7)}(Pool(x)))) \\
X_3 = Conv_{(1\times1)}(ReLU(Conv_{(1\times1)}(AvgPool(Conv_{(3\times3)}(x))))) \\
X_4 = Conv_{(1\times1)}(ReLU(Conv_{(1\times1)}(MaxPool(Conv_{(3\times3)}(x))))) \\
Y_2 = x + (X_2 \otimes (X_3 + X_4))
\end{gather}
where $x$ denotes the input of the MSAU and $Y_2$ represents the output feature map of the MSAU. 
Among the formulas, $Conv_{k\times k}(\cdot)$ are normal convolution operation. $\otimes$ denotes element-by-element multiplication, $Pool(\cdot)$ denotes the adaptive average pooling, $AvgPool(\cdot)$ is average pooling, $MaxPool(\cdot)$ is maximum pooling, $ReLU(\cdot)$ is rectified linear unit and $Sigmoid(\cdot)$ is the Sigmoid activation function.

\subsection{Feature Fusion Module}

Motivated by by the effectiveness  of Mamba in linear-complexity sequence modeling, we design a Feature Fusion Module (FFM) by introduce 2D-Selective-Scan (SS2D) block for better capturing global representations with less network parameters and computational quantities.
The FFM enriches the feature diversity by integrating different scale feature information from the multi-level the MSAU and the encoder through the concatenation operation.
Then, the SS2D block further extracts and fuses the features through a series of linear transformations and 2D convolution operations, which employs a selective scanning mechanism to enhance the feature representation ability.
Finally, Feed-Forward Network (FFN) performs a nonlinear transformation to adjust the weight distribution of features, highlighting the key features and suppressing the redundant information, so as to improve performance in handling complex tasks.
The designed FFM can effectively fuse multi-scale features and capture both local detail information and overall semantic features, great improving the performance of the model in semantic segmentation tasks.
The complete operation is shown as follows:
\begin{gather}
    X_{FFN} = FFN(SS2D(Concat(x_{encoder},x_{MSAU1},x_{MSAU2})))\\
    Y_3 = X_{FFN} + x_{encoder}
\end{gather}
where $x_{encoder},x_{MSAU1},x_{MSAU2}$ denotes the out of the Encoder and MSAU resectively, $Y_3$ denotes the output feature map of the FFM. 
Among the formulas, $Concat(\cdot)$ is normal concatenation operation. $SS2D(\cdot)$ is the 2D-Selective-Scan block and $FFN(\cdot)$ is the eed-Forward Network.

\section{Experiments}\label{sec4}
\subsection{Datasets}
\begin{itemize}
    \item \textbf{Cityscapes.} This dataset is composed of high-quality 5,000 images, annotated at the pixel level. The images are primarily scenes of driving within urban settings, captured across 50 different cities with a resolution of 2,048×1,024. The dataset was divided into training sets(2,975 images), validation sets (500 images), and test sets (1,525 images)

    \item \textbf{CamVid.} The CamVid dataset, developed by the University of Cambridge, contains urban road scene images captured from a driving perspective (960×720 resolution). Its 700+ annotated samples support supervised learning, featuring 11 representative object classes that effectively capture urban road elements. This diversity in objects and well-annotated classes makes it particularly suitable for our segmentation accuracy research.
\end{itemize}

\begin{figure}[h]
    \centering
    \includegraphics[width=0.6\linewidth]{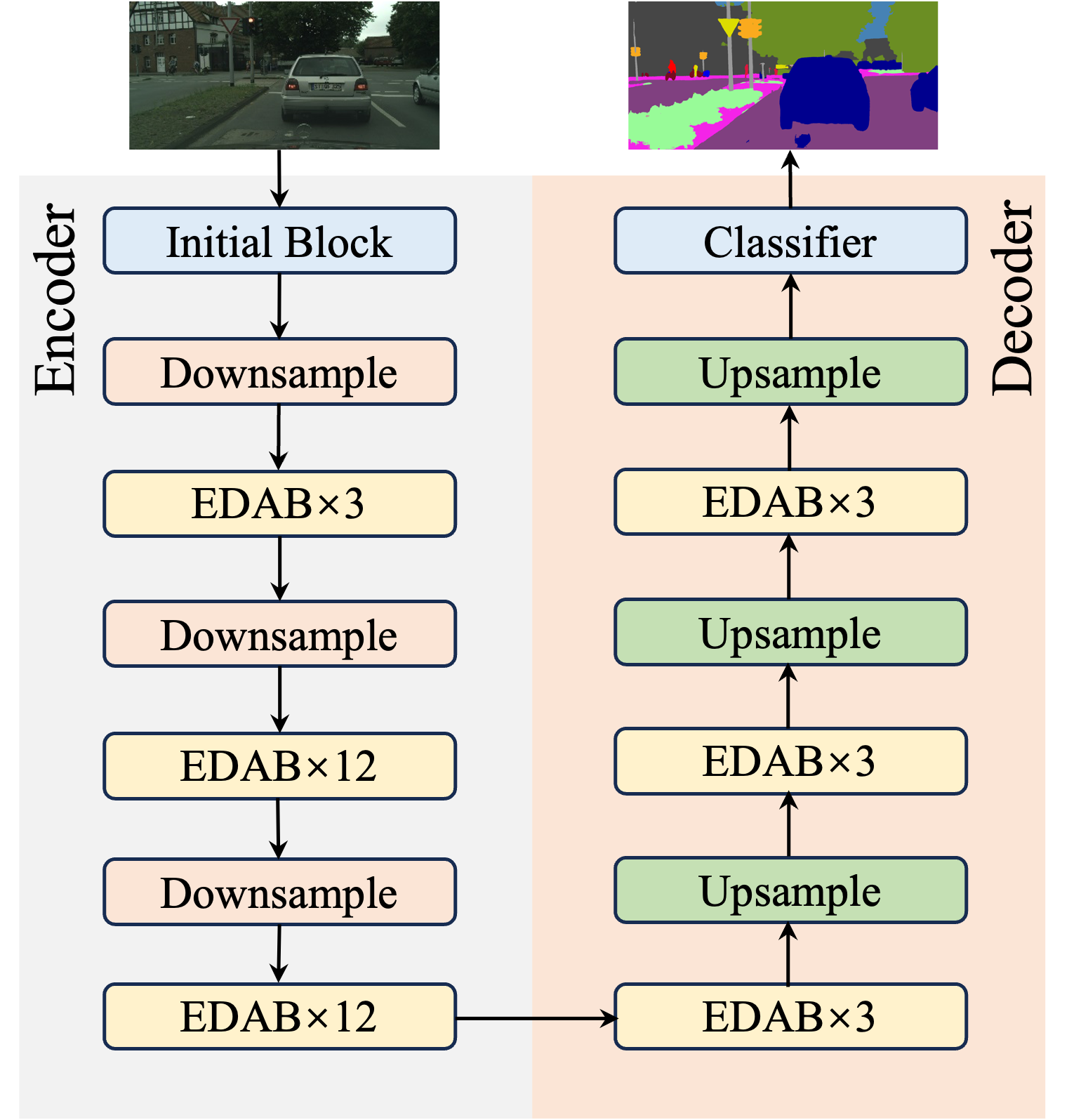}
    \caption{The simple structure of the baseline model}
    \label{fig:baseline}
\end{figure}

\subsection{Ablation Studies}

We design a series of ablation experiments to
validate the effectiveness of each module in our proposed model.
As shown in Figure \ref{fig:baseline}, The baseline model used for comparison is structured as simple U-shape type, including the standard Encoder and Decoder. 
The Encoder and Decoder consist of multiple lightweight enhanced dual-attention blocks (EDABs), which are modeled to achieve an average mIoU of 69.92\% on the Camvid validation set.

\begin{table}[h]
\caption{Extensive ablation study for the proposed ECMNet on Camvid dataset.}\label{tab：Ablation study for the proposed ECMNet on Camvid dataset}%
\setlength{\tabcolsep}{2pt}
\begin{tabular}{ccccccccccc}
\toprule
 \multirow{3}{*}{\textbf{Architecture}}
 &\multicolumn{7}{@{}c@{}}{\textbf{Method}}&\multirow{3}{*}{\textbf{Parameter (K)$\downarrow$ }}& \multirow{3}{*}{\textbf{FLOPs (G)$\downarrow$}} & \multirow{3}{*}{\textbf{mIoU (\%)$\uparrow$} }\\\cmidrule{2-8}
  &\multicolumn{3}{c}{\textbf{Long Connection}}& \multicolumn{3}{c}{\textbf{MSAU}}&\multirow{2}{*}{\textbf{FFM}}&&&\\\cmidrule{2-7}
  &1&2&3&1&2&3&&&&\\
 \hline
 \midrule

\textbf{Baseline} &-&-&-&-&-&-&-& 775.57  & 7.56 & 69.92 \\
\textbf{A1}& \checkmark &-&-&-&-&-&-& 777.93 & 7.57 & 70.53\textsuperscript{\textcolor{blue}{0.61$\uparrow$}}\\
\textbf{A2}&\checkmark &\checkmark&-&-&-&-&-&796.41&7.64&70.92\textsuperscript{\textcolor{blue}{1.00$\uparrow$}}\\
\textbf{A3}&\checkmark &\checkmark&\checkmark&-&-&-&-&805.67&7.79&71.21\textsuperscript{\textcolor{blue}{1.29$\uparrow$}}\\
\textbf{B1}&-&-&-& \checkmark &-&-&-& 787.34&7.57&71.45\textsuperscript{\textcolor{blue}{1.53$\uparrow$}}\\
\textbf{B2}&-&-&-& \checkmark &\checkmark&-&-& 805.82&7.67&72.65\textsuperscript{\textcolor{blue}{2.73$\uparrow$}}\\
\textbf{B3}&-&-&-& \checkmark &\checkmark&\checkmark&-& 815.08&7.90&73.22\textsuperscript{\textcolor{blue}{3.30$\uparrow$}}\\
\textbf{C1}&-&-&-&-&-&-&\checkmark&827.80&7.80&70.75\textsuperscript{\textcolor{blue}{0.83$\uparrow$}}\\
\textbf{C2}&\checkmark &\checkmark&\checkmark&-&-&-&\checkmark&863.93&8.06&71.03\textsuperscript{\textcolor{blue}{1.11$\uparrow$}}\\
\textbf{C3 (ours)}&\checkmark &\checkmark&\checkmark&\checkmark&\checkmark&\checkmark&\checkmark&871.11&8.27&\textbf{73.62}\textsuperscript{\textcolor{blue}{3.70$\uparrow$}}\\
\botrule
\end{tabular}
\footnotetext{A, B, C denote the long connection, the feature enhancement and the feature fusion respectively. No. of A, B and C denotes the the stack of same or different modules.}
\end{table}

In the long connection ablation experiments (A Group), the effect of gradually adding Line 1, Line 2 , and Line 3 is investigated. 
The observed 0.61\% enhancement after adding Line 1 substantiates that shallow information effectively aids semantic feature information reconstruction.
Meanwhile, With three long skip connections, the model achieves a 1.29\% mIoU enhancement. 
These results further demonstrate the significance of long-range skip connections for semantic segmentation task.
In the MSAU ablation experiments (B group), the MSAU module is added gradually in the long connection. 
A comparison between B1 and A1 reveals that adding the MSAU module to long connections only adds 9.43K parameters, but improves the  performance by 0.92\% of mIoU.
In the last ablation experiments (C Group), the introduction of the Feature Fusion Module (FFM) improves the performance of the model by 1.11\% of mIoU.
Finally, as the finalized architecture (C3), our proposed ECMNet improves performance by 3.7\% mIoU compared to the baseline model.
All the above experiments shown in Table \ref{tab：Ablation study for the proposed ECMNet on Camvid dataset} fully validate the efficacy of our proposed modules and strategies


\begin{table}[h]
\centering
\caption{Performance comparison of our proposed ECMNet and other representative methods on the Cityscapes dataset.}
\label{tab: comparison on the Cityscapes dataset}
\setlength{\tabcolsep}{5pt}
\begin{tabular}{lccccccc}
\toprule
\multirow{2}{*}{Method} & \multirow{2}{*}{Year} & Resolution & \multirow{2}{*}{Backbone} & {Parameter} & {FLOPs} & {Speed} & {mIoU} \\
 & & (\text{width$\times$height}) & & {(M)$\downarrow$} & {(G)$\downarrow$} & {(FPS)$\uparrow$} & {(\%)$\uparrow$} \\
\midrule
\hline
SegNet\cite{badrinarayanan2017segnet} & 2017 & 640$\times$360 & VGG-16 & 29.50 & 286.0 & 17 & 57.0 \\
ENet\cite{paszke2016enet} & 2016 & 512$\times$1024 & No & \colorbox{lightgray}{0.36} & 3.8 & 135 & 58.3 \\
ESPNet\cite{mehta2018espnet} & 2018 & 512$\times$1024 & ESPNet & \colorbox{lightgray}{0.36} & -- & 113 & 60.3 \\
NDNet\cite{yang2020ndnet} & 2021 & 512$\times$1024 & No & 0.50 & 3.5 & 120 & 61.1 \\
CGNet\cite{wu2020cgnet} & 2021 & 360$\times$640 & No & 0.49 & 6.0 & -- & 64.8 \\
ADSCNet\cite{wang2020adscnet} & 2020 & 512$\times$1024 & No & -- & -- & 77 & 67.5 \\
ERFNet\cite{romera2017erfnet} & 2017 & 512$\times$1024 & No & 2.10 & -- & 42 & 68.0 \\
BiseNet-v1\cite{zhao2022real} & 2018 & 768$\times$1536 & Xception & 5.80 & 14.8 & 106 & 68.4 \\
ICNet\cite{zhao2018icnet} & 2018 & 1024$\times$1024 & PSPNet-50 & 26.50 & 28.3 & 30 & 69.5 \\
DABNet\cite{li2019dabnet} & 2019 & 1024$\times$2048 & No & 0.76 & 42.4 & 28 & 70.1 \\
CFPNet\cite{ding2024cfpnet} & 2021 & 1045$\times$2048 & No & 0.55 & -- & 30 & 70.1 \\
FPENet\cite{liu2019feature} & 2019 & 512$\times$1024 & No & 0.40 & 12.8 & 55 & 70.1 \\
LEDNet\cite{wang2019lednet} & 2019 & 512$\times$1024 & No & 0.94 & -- & 40 & 70.6 \\
DFANet\cite{li2019dfanet} & 2019 & 1024$\times$1024 & Xception & 7.80 & \colorbox{lightgray}{3.4} & 100 & 71.3 \\
STDC1-50\cite{fan2021rethinking} & 2021 & 512$\times$1024 & -- & 8.40 & -- & 87 & 71.9 \\
SegFormer\cite{xie2021segformer}  & 2021 & 512$\times$1024 & MiT-B0 & 3.80 & 17.7 & 48 & 71.9 \\
MSCFNet\cite{gao2021mscfnet} & 2022 & 512$\times$1024 & No & 1.15 & 17.1 & 50 & 71.9 \\
FPANet\cite{wu2022fpanet} & 2022 & 512$\times$1024 & -- & 14.10 & -- & -- & 72.0 \\
MLFNet\cite{fan2022mlfnet} & 2023 & 512$\times$1024 & ResNet-34 & 13.03 & 15.5 & 72 & 72.1 \\
BiseNet-v2\cite{yu2021bisenet} & 2021 & 512$\times$1024 & Xception & 3.40 & 21.2 & 156 & 72.6 \\
MGSeg\cite{he2021mgseg} & 2021 & 1024$\times$1024 & ShuffleNet-v2 & 4.50 & 16.2 & 101 & 72.7 \\
PCNet\cite{lv2021parallel} & 2022 & 1024$\times$2048 & Scratch & 3.40 & 21.2 & 156 & 72.6 \\
LETNet\cite{xu2023lightweight} & 2023 & 512$\times$1024 & No & 0.95 & 13.6 & 150 & 72.8 \\
SCTNet-S\cite{xu2024sctnet} & 2024 & 512$\times$1024 & No & 4.6 & 451.2 & \colorbox{lightgray}{160.3} & 72.8 \\
HSB-Net\cite{li2021hierarchical} & 2021 & 512$\times$1024 & ResNet-34 & 12.10 & -- & 124 & 73.1 \\
LBN-AA\cite{dong2020real} & 2021 & 448$\times$896 & No & 6.20 & 49.5 & 51 & \colorbox{lightgray}{73.6} \\
\midrule
\hline
\textbf{ECMNet (Ours)} & -- & 1024$\times$1024 & No & 0.87 & 8.27 & 43 & \textbf{70.6} \\
\bottomrule
\end{tabular}
\footnotetext{The gray box denotes the best value of the current metric.}
\end{table}

\subsection{Comparisons with SOTA Methods}

In this section, we compare state-of-the-art semantic segmentation methods in recent years on the Cityscapes and CamVid datasets to verify that our approach achieves a better balance between performance and parameters.
Our evaluation is based on three key metrics:
model parameters, floating-point
operations (FLOPs) and mIoU.

\textbf{Evaluation Results on Cityscapes Dataset.} 
As shown in Table \ref{tab: comparison on the Cityscapes dataset}, the model with a larger number of parameters and computation obviously achieves excellent segmentation results.
However, computational complexity of model is high and its operation speed is slow, which is unsuitable for real-time intelligent embedded devices.
In contrast, lightweight models like NDNet\cite{yang2020ndnet}, CGNet\cite{wu2020cgnet}, CFPNet\cite{ding2024cfpnet} , LEDNet\cite{wang2019lednet} and LETNet\cite{xu2023lightweight} are computationally efficient, which lack overall performance, especially in accuracy. 
Obviously, the LBN-AA\cite{dong2020real} achieved the highest mIoU with 6.2M model parameters which are far more than our proposed approach.
Meanwhile, the ESPNet\cite{mehta2018espnet} utilized the least parameters to realize 60.3\% mIoU, which is significantly lower than the performance of our method.
Our proposed ECMNet only with a 0.87M parameters achieved a higher 70.6\% mIoU.
Our proposed ECMNet can get better segmentation results with less model parameters, which benefits from well-design structure, and the utilization of the Mamba. 
These results fully demonstrate that our proposed model can achieve a excellent balance between model parameters and performance.

\textbf{Evaluation Results on CamVid Dataset.} 
As shown in Table \ref{tab:comparison on the CamVid dataset}, to further verify the effectiveness and generalization capacity of our proposed ECMNet, we conducted comparative experments with our method and other lightweight models on the CamVid dataset. 
Obviously, the MGSeg\cite{he2021mgseg} just achieved the 72.7\% mIoU with 13.3M model parameters which is lower performance and larger model parameters compared our proposed method.
Therefore, our method has achieved the best accuracy by only using 0.87M parameters.
Compared to Cityscapes, the higher overall performance on the CamVid
dataset is due to our designed modules and strategies, which better capture feature of small size datasets. 
Per-class results are detailed in Table \ref{tab:per-class results on CamVid} further demonstrate the advantages of our proposed ECMNet.

\begin{table}[h]
\centering
\caption{Performance comparison  of our proposed ECMNet and other representative methods on the CamVid dataset.}
\label{tab:comparison on the CamVid dataset}
\begin{tabular}{lcccccc}
\toprule
\multirow{2}{*}{Method} & \multirow{2}{*}{Year} & \multirow{2}{*}{Resolution} & \multirow{2}{*}{Backbone} & {Parameter} & {Speed} & {mIoU} \\
 & & & & {(M)$\downarrow$} & {(FPS)$\downarrow$} & {(\%)$\uparrow$} \\
\midrule
\hline
ENet\cite{paszke2016enet} & 2016 & $360\times480$ & No & \colorbox{lightgray}{0.36}& 68 & 51.3 \\
SegNet\cite{badrinarayanan2017segnet} & 2017 & $360\times480$ & VGG-16 & 29.45 & 87 & 55.6 \\
NDNet\cite{yang2020ndnet} & 2021 & $360\times480$ & No & 0.50 & 78 & 57.2 \\
DFANet\cite{li2019dfanet} & 2019 & $360\times480$ & Xception & 7.80 & 120 & 64.7 \\
DABNet\cite{li2019dabnet} & 2019 & $360\times480$ & No & 0.76 & \colorbox{lightgray}{136} & 66.4 \\
FDDWNet\cite{liu2020fddwnet} & 2020 & $360\times480$ & No & 0.80 & 79 & 66.9 \\
ICNet\cite{zhao2018icnet} & 2018 & $720\times960$ & PSPNet-50 & 26.50 & 28 & 67.1 \\
FBSNet\cite{gao2022fbsnet} & 2023 & $360\times480$ & No & 0.62 & 120 & 68.9 \\
MSCFNet\cite{gao2021mscfnet} & 2022 & $360\times480$ & No & 1.15 & 110 & 69.3 \\
LETNet\cite{xu2023lightweight} & 2023 & $360\times480$ & No & 0.95 & 21 & 70.5 \\
HAFormer\cite{xu2024haformer} & 2024 & $360\times480$ & No & 0.60 & 118 & 71.1 \\
MGSeg\cite{he2021mgseg} & 2021 & $736\times736$ & ResNet-18 & 13.3 & 127 & 72.7 \\ 
\midrule
\hline
\textbf{ECMNet (Ours)} & - & $360\times480$ & No & 0.87 & 55 & \colorbox{lightgray}{\textbf{73.6}} \\
\bottomrule
\end{tabular}
\footnotetext{The gray box denotes the best value of the current metric.}
\end{table}

\section{Conclusion}\label{sec13}

In this study, we proposed a lightweight semantic segmentation network that combines Mamba and Convolutional Neural Networks (CNNs). 
We fused the local feature extraction capability of convolutional neural networks with long-range dependencies of Mamba to model. 
Specifically, we introduced a Feature Fusion Module (FFM) as a capsule-based framework in the middle of the model, which can better capture global feature information.
Additionally, an Enhanced Dual Attention Module (EDAB) designed in the convolutional neural network learned more local feature information while ensuring simplicity and lightweight. 
Meanwhile, in order to compensate for the local feature information lost by CNNs, multi-scale long connections are utilized in the model. 
Moreover, We design a Multi-Scale Attention Unit (MSAU) for cross-layer connections, effectively boosting discriminative features and attenuating noise. 
Extensive experimental results demonstrate that our proposed model achieves an excellent balance between model scale and performance.

\begin{sidewaystable}
\centering
\caption{Performance comparison of our proposed ECMNet and the state-of-arts lightweight methods about per-class results on the CamVid dataset.}
\label{tab:per-class results on CamVid}
\setlength{\tabcolsep}{2pt}
\begin{tabular}{lcccccccccccccccccccc}
\toprule
{Method} 
 & {Roa} & {Sid} & {Bui} & {Wal} & {Fen} & {Pol} & {TLi} & {TSi} & {Veg} & {Ter} & {Sky} & {Ped} & {Rid} & {Car} & {Tru} & {Bus} & {Tra} & {Mot} & {Bic} & {mIoU(\%)} \\
\midrule
\hline
Enet\cite{paszke2016enet} & 96.3 & 74.2 & 75.0 & 32.2 & 33.2 & 43.4 & 34.1 & 44.0 & 88.6 & 61.4 & 90.6 & 65.5 & 38.4 & 90.6 & 36.9 & 50.5 & 48.1 & 38.8 & 55.4 & 58.3 \\
ESPNet\cite{mehta2018espnet} & 97.0 & 77.5 & 76.2 & 35.0 & 36.1 & 45.0 & 35.6 & 46.3 & 90.8 & 63.2 & 92.6 & 67.0 & 40.9 & 92.3 & 38.1 & 52.5 & 50.1 & 41.8 & 57.2 & 60.3 \\
CGNet\cite{wu2020cgnet} & 95.5 & 78.7 & 88.1 & 40.0 & 43.0 & 54.1 & 59.8 & 63.9 & 89.6 & 67.6 & 92.9 & 74.9 & 54.9 & 90.2 & 44.1 & 59.5 & 25.2 & 47.3 & 60.2 & 64.8 \\
ESPNet-v2\cite{lin2023lightweight} & 97.3 & 78.6 & 88.8 & 43.5 & 42.1 & 49.3 & 52.6 & 60.0 & 90.5 & 66.8 & 93.3 & 72.9 & 53.1 & 91.8 & 53.0 & 65.9 & 53.2 & 44.2 & 59.9 & 66.2 \\
ERFNet\cite{romera2017erfnet} & 97.7 & 81.0 & 89.8 & 42.5 & 48.0 & 56.3 & 59.8 & 65.3 & 91.4 & 68.2 & 94.2 & 76.8 & 57.1 & 92.8 & 50.8 & 60.1 & 51.8 & 47.3 & 61.7 & 68.0 \\
DABNet\cite{li2019dabnet} & 96.8 & 78.5 & 90.9 & 45.4 & 50.2 & 59.1 & 65.2 & 70.8 & 92.5 & 68.2 & 94.6 & 80.5 & 58.5 & 92.7 & 52.7 & 67.2 & 50.9 & 50.4 & 65.7 & 70.0 \\
CFPNet\cite{ding2024cfpnet} & 97.8 & 81.4 & 90.5 & 46.4 & 50.6 & 56.4 & 61.5 & 67.7 & 92.1 & 68.9 & 94.3 & 80.4 & 60.7 & 93.9 & 51.4 & 68.0 & 50.8 & 51.2 & 67.7 & 70.1 \\
LEDNet\cite{wang2019lednet} & \colorbox{lightgray}{98.1} & 79.5 & \colorbox{lightgray}{91.6} & \colorbox{lightgray}{47.7} & 49.9 & \colorbox{lightgray}{62.8} & 61.3 & \colorbox{lightgray}{72.8} & \colorbox{lightgray}{92.6} & \colorbox{lightgray}{61.2} & \colorbox{lightgray}{94.9} & \colorbox{lightgray}{76.2} & \colorbox{lightgray}{53.7} & 90.9 & 64.4 & 64.0 & 52.7 & \colorbox{lightgray}{44.4} & \colorbox{lightgray}{71.6} & 70.6 \\
\midrule
\hline
\textbf{ECMNet (Ours)} & 97.1 & \colorbox{lightgray}{80.8} & 90.9 & 44.2 & \colorbox{lightgray}{53.4} & 60.8 & \colorbox{lightgray}{61.9} & 72.4 & 91.7 & 60.4 & 93.9 & 75.2 & 52.7 & \colorbox{lightgray}{92.0} & \colorbox{lightgray}{65.3} & \colorbox{lightgray}{76.5} & \colorbox{lightgray}{66.5} & 37.8 & 69.1 & \colorbox{lightgray}{\textbf{70.6}} \\
\bottomrule
\end{tabular}
\footnotetext{The gray box denotes the best mIoU of the current class. Roa, Sid, Bui, Wal, Fen, Pol, TLi, TSi, Veg, Ter, Sky, Ped, Rid, Car, Tru, Mot and Bic reprsent Road, Sidewalk, Building, Wall, Fence, Pole, Traffic Light, Traffic Sign, Vegtation, Terrain, Sky, Pedestrain, Rider, Car, Truck, Motorcycle and Bicycle respectively.
}
\end{sidewaystable}




\clearpage
\bibliography{sn-bibliography}

\end{document}